\documentclass[twoside]{article}

\usepackage{
    amsfonts,
    amsthm,
    graphicx,
    url,
    xcolor,
    xfrac,
}
\usepackage[accepted]{aistats2023}
\usepackage[round]{natbib}
\usepackage[font=small,labelfont=bf,labelsep=quad,justification=justified,singlelinecheck=false]{caption}
\usepackage[font=small,labelfont=bf,labelsep=quad,singlelinecheck=true]{subcaption}
\usepackage[hyperfootnotes=false]{hyperref}
\usepackage[capitalise,nameinlink,noabbrev]{cleveref}

\newtheoremstyle{example}  
    {10pt}  
    {10pt}  
    {\itshape}  
    {}  
    {\bfseries}  
    {\quad}  
    { }  
    {}  
\theoremstyle{example}
\newtheorem{example}{Example}

\hypersetup{
    colorlinks,
    linkcolor={blue!50!black},
    citecolor={blue!50!black},
    urlcolor={blue!50!black},
}

\newcommand{\email}[1]{\href{mailto:#1}{\nolinkurl{#1}}}

\newcommand{\shorturl}[1]{\href{https://#1}{\nolinkurl{#1}}}

\creflabelformat{equation}{#2\textup{#1}#3}

\newlength{\figureheight}
\newcommand{\entropy}[1]{\mathrm{H}\!\left[#1\right]}
\newcommand{\mutualinfo}[1]{\mathrm{I}\!\left(#1\right)}
\newcommand{\expectation}[2]{\mathbb{E}_{#1}\!\left[#2\right]}
\newcommand{\kldivergence}[2]{\mathrm{KL}\!\left[#1\,\|\,#2\right]}

\newcommand{\basedata}[0]{\mathcal{D}_{\mathrm{base}}}
\newcommand{\pooldata}[0]{\mathcal{D}_{\mathrm{pool}}}
\newcommand{\traindata}[0]{\mathcal{D}_{\mathrm{train}}}
\newcommand{\testdata}[0]{\mathcal{D}_{\mathrm{test}}}

\newcommand{\ppool}[0]{p_\mathrm{pool}}
\newcommand{\ptarg}[0]{p_*}

\begin{document}

\raggedbottom

\runningauthor{Freddie Bickford Smith*, Andreas Kirsch*, Sebastian Farquhar, Yarin Gal, Adam Foster, Tom Rainforth}

\twocolumn[
    \aistatstitle{Prediction-Oriented Bayesian Active Learning}
    \aistatsauthor{Freddie Bickford Smith* \And Andreas Kirsch* \And Sebastian Farquhar}
    \aistatsaddress{University of Oxford \And University of Oxford \And University of Oxford}
    \aistatsauthor{Yarin Gal \And Adam Foster \And Tom Rainforth}
    \aistatsaddress{University of Oxford \And Microsoft Research \And University of Oxford}
]

\begin{abstract}

    Information-theoretic approaches to active learning have traditionally focused on maximising the information gathered about the model parameters, most commonly by optimising the BALD score.
    We highlight that this can be suboptimal from the perspective of predictive performance.
    For example, BALD lacks a notion of an input distribution and so is prone to prioritise data of limited relevance.
    To address this we propose the expected predictive information gain (EPIG), an acquisition function that measures information gain in the space of predictions rather than parameters.
    We find that using EPIG leads to stronger predictive performance compared with BALD across a range of datasets and models, and thus provides an appealing drop-in replacement.

\end{abstract}
\section{Introduction}

Active learning \citep{angluin1988queries,atlas1989training,liu2022survey,settles2012active} allows us to make the most of limited label budgets by adaptively deciding which inputs to acquire labels for when training a model.
A principled basis for acquisition is to formalise a label's utility through the information it provides.
Doing this requires a probabilistic generative model for possible future labels, leading to an approach known as Bayesian active learning \citep{gal2017deep,houlsby2011bayesian,mackay1992evidence,mackay1992information}.

Historically the literature has focused on trying to maximise the expected information gain (EIG) in the model parameters.
This yields an acquisition function typically known as the BALD score, having been popularised by a method called Bayesian active learning by disagreement \citep{houlsby2011bayesian}.
It has been successfully applied in a number of settings, including computer vision \citep{gal2017deep} and natural-language processing \citep{shen2018deep}.

In this work we highlight that BALD can be misaligned with our typical overarching goal of making effective predictions on unseen inputs.
Specifically it targets information about the model parameters, but not all such information is equally useful when it comes to making predictions.
With a nonparametric model, for instance, we can gain an infinite amount of information about the model parameters without any of it being relevant to prediction on inputs of interest.
In short, BALD lacks a notion of how the model will be used and so fails to ensure that the data acquired is relevant to our particular predictive task.

This has considerable practical implications.
Real-world datasets are often messy, with inputs that vary widely in their relevance to a given task.
Large pools of audio, images and text commonly fit this description \citep{ardila2020common,gemmeke2017audio,mahajan2018exploring,radford2021learning,raffel2020exploring,sun2017revisiting}.
We show that BALD can be actively counterproductive in cases like these, picking out the most obscure, least relevant inputs.

\begin{figure*}[t]
    \centering
    \includegraphics[width=\linewidth,trim={0.2cm 0 0.2cm 0}]{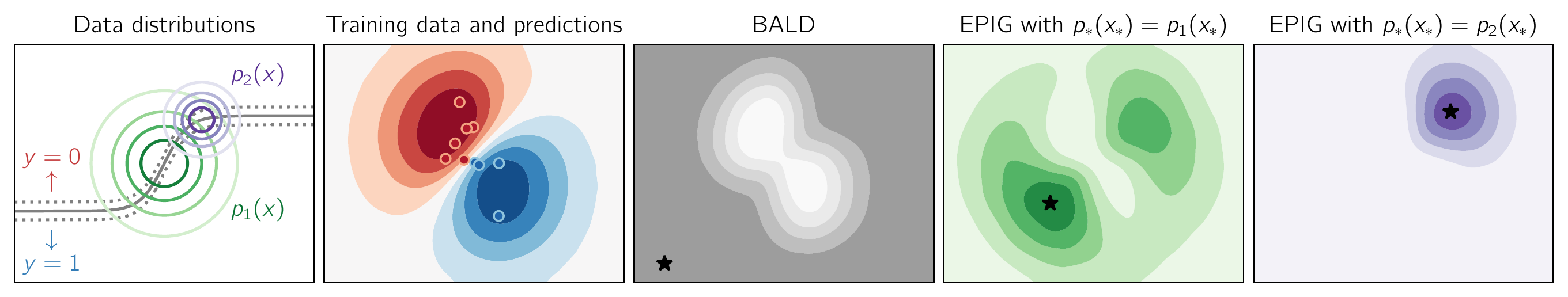}
    \caption{
        The expected predictive information gain (EPIG) can differ dramatically from the expected information gain in the model parameters (BALD).
        BALD increases (darker shading) as we move away from the existing data, yielding a distant acquisition (star) when maximised.
        It seeks a global reduction in parameter uncertainty, regardless of any input distribution.
        In contrast EPIG is maximised only in regions of relatively high density under the target input distribution, $\ptarg(x_*)$.
        It seeks a reduction in parameter uncertainty only insofar as it reduces predictive uncertainty on samples from $\ptarg(x_*)$.
        See \Cref{sec:2d_data} for details.
    }
    \label{fig:heatmaps}
\end{figure*}

To address BALD's shortcomings we propose the expected predictive information gain (EPIG), an alternative acquisition function.
We derive EPIG by returning to the foundational framework of Bayesian experimental design \citep{lindley1956measure}, from which BALD itself is derived.
Whereas BALD is the EIG in the model parameters, EPIG is the EIG in the model's predictions: it measures how much information the label of a candidate input is expected to provide about the label of a random target input.
While BALD favours global reductions in parameter uncertainty, EPIG favours only information that reduces downstream predictive uncertainty (\Cref{fig:heatmaps}).
Thus EPIG allows us to directly seek improvements in predictive performance.

The randomness of the target input in EPIG is critical.
We do not aim for predictive information gain on a particular input or set of inputs.
Instead the gain is in expectation with respect to a target input distribution.
This can be chosen to be the same distribution that the pool of unlabelled inputs is drawn from, or it can be a distinct distribution that reflects a downstream task of interest.

We find that EPIG often produces notable gains in final predictive performance over BALD across a range of datasets and models.
EPIG's gains are largest when the pool of unlabelled inputs contains a high proportion of irrelevant inputs with respect to the target input distribution.
But its advantage still holds when the pool is directly drawn from this distribution.
As such, it can provide a simple and effective drop-in replacement for BALD in many settings.
\section{Background}
\label{sec:background}

We consider supervised learning of a probabilistic predictive model, $p_{\phi}(y|x)$, where $x\in\mathcal{X}$ is an input, $y\in\mathcal{Y}$ is a label and $\phi$ indexes the set of models we can learn.
We assume the model has some underlying stochastic parameters, $\theta$, such that
we can write
\begin{align}
    p_{\phi}(y|x)             & = \expectation{p_{\phi}(\theta)}{p_{\phi}(y|x,\theta)}
    \label{eq:marg_pred_posterior}
    \\
    p_{\phi}(y_1,y_2|x_1,x_2) & = \expectation{p_{\phi}(\theta)}{p_{\phi}(y_1,y_2|x_1,x_2,\theta)}
    .
    \label{eq:joint_pred_posterior}
\end{align}
We also assume predictions are independent given $\theta$, which gives $p_{\phi}(y_1,y_2|x_1,x_2,\theta) = p_{\phi}(y_1|x_1,\theta)p_{\phi}(y_2|x_2,\theta)$.

The class of models satisfying our assumptions is broad.
It includes effectively all Bayesian models, for which $p_{\phi}(y|x,\theta)$ is a fixed likelihood function and $p_{\phi}(\theta)\!=\!p(\theta|\mathcal{D})$ is a posterior given observed data $\mathcal{D}$.
Also included are ensembles \citep{dietterich2000ensemble} and neural networks with stochasticity in a subset of parameters \citep{sharma2023bayesian}.

\subsection{Active learning}
\label{sec:active_learning}

\begin{figure}[t]
    \centering
    \includegraphics[height=\figureheight,trim={0 5cm 0 5cm}]{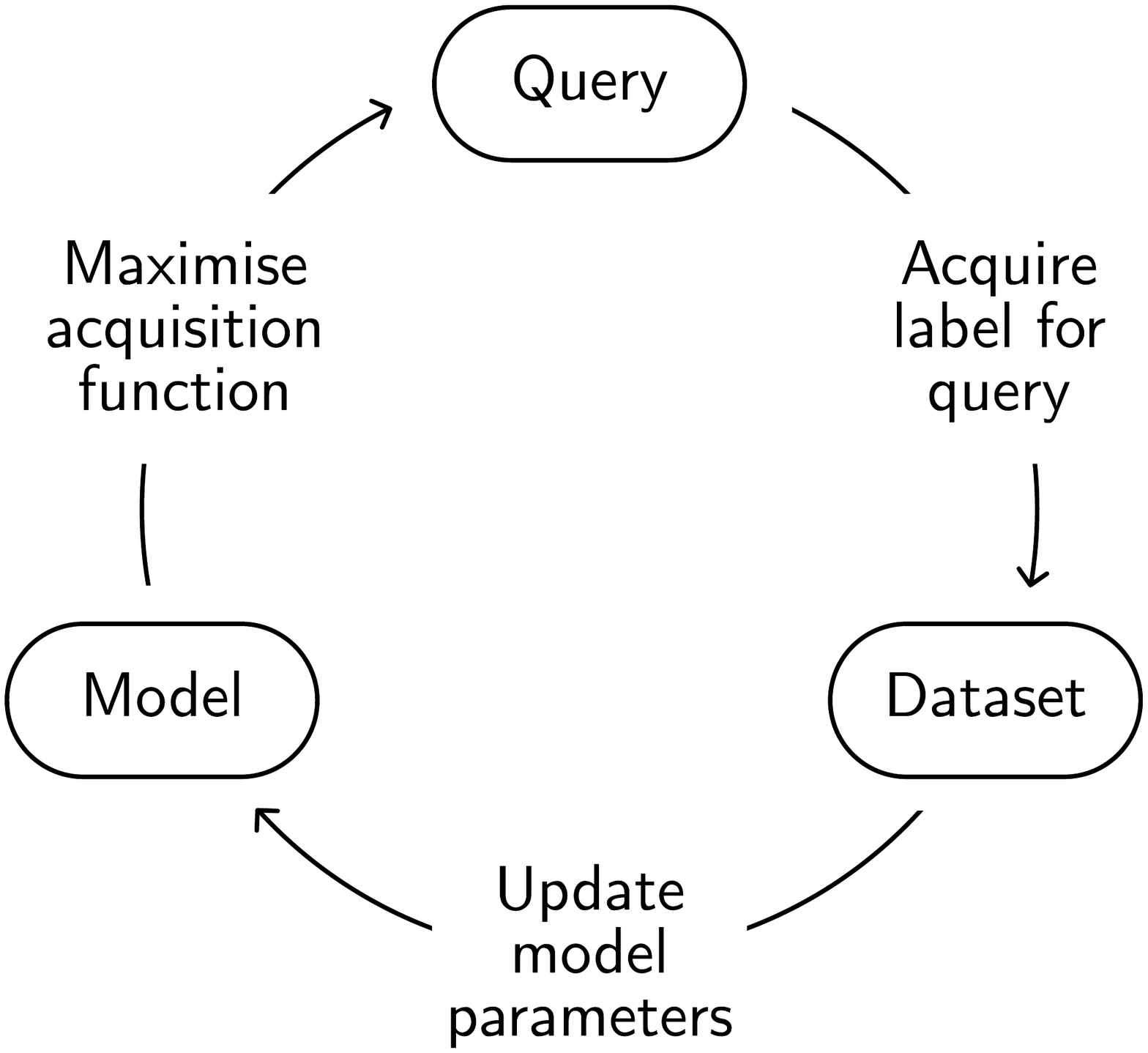}
    \caption{
        Active learning typically loops over selecting a query, acquiring a label and updating the model parameters.
        In this work we focus on the acquisition function used to select queries.
        We consider pool-based settings, where the acquisition function is maximised across a fixed, finite set of unlabelled inputs.
    }
    \label{fig:active_learning_loop}
\end{figure}

In the supervised setting, active learning involves having an algorithm select which labels to acquire when training a model \citep{settles2012active}.
Typically acquisition takes place across a number of steps.
Each step, $t$, consists of three parts (\Cref{fig:active_learning_loop}).
First, the algorithm selects a query input, $x_t$, to acquire a label for---or sometimes a batch of inputs \citep{kirsch2019batchbald}.
It does this by maximising an acquisition function, which is intended to capture the expected utility of acquiring the label for a given input.
Often the set of candidate inputs is a fixed, finite pool, $\pooldata = \{x_i\}_{i=1}^{N}$.
We focus on such settings, known as pool-based active learning \citep{lewis1994sequential}.
Second, the algorithm samples a label, $y_t$, from the true conditional label distribution, $p(y|x=x_t)$, and incorporates $(x_t,y_t)$ into the training dataset.
Third, the predictive model, $p_{\phi}(y|x)$, is updated.

\subsection{Bayesian experimental design}

Bayesian experimental design \citep{chaloner1995bayesian,lindley1956measure,rainforth2023modern} is a formal framework for quantifying the information gain from an experiment.
In the context of active learning we can view the input, $x$, as the design of the experiment and the acquired label, $y$, as the outcome of the experiment.

Let $\psi$ be a quantity we are aiming to learn about.
Given a prior, $p(\psi)$, and a likelihood function, $p(y|x,\psi)$, both of which could be implicit, we can quantify the information gain in $\psi$ due to an experiment, $(x,y)$, as the reduction in Shannon entropy in $\psi$ that results from observing $(x,y)$:
\begin{align*}
    \mathrm{IG} & _{\psi}(x,y) = \entropy{p(\psi)} - \entropy{p(\psi|x,y)}                              \\
                & = \expectation{p(\psi)}{-\log p(\psi)} - \expectation{p(\psi|x,y)}{-\log p(\psi|x,y)}
    ,
\end{align*}
where $p(\psi|x,y) \propto p(\psi)p(y|x,\psi)$ is the posterior that results from a Bayesian update on observing $(x,y)$.

Since $y$ is a random variable, we consider the expected information gain (EIG) across possible realisations of $y$, simulating outcomes using $p_\psi(y|x)=\expectation{p(\psi)}{p(y|x,\psi)}$, the marginal predictive distribution:
\begin{align*}
    \mathrm{EIG}_{\psi}(x) = \expectation{p_\psi(y|x)}{\entropy{p(\psi)} - \entropy{p(\psi|x,y)}}
    .
\end{align*}
This is the expected reduction in uncertainty in $\psi$ after conditioning on $(x,y)$.
Equivalently it is $\mutualinfo{\psi;y|x}$, the mutual information between $\psi$ and $y$ given $x$.

\subsection{Bayesian active learning by disagreement (BALD)}

Bayesian active learning has traditionally targeted information gain in the model parameters, setting $\psi$ to $\theta$.
This yields what is often referred to in the active-learning literature as the BALD score \citep{houlsby2011bayesian}:
\begin{align*}
    \mathrm{BALD}(x) & = \expectation{p_{\phi}(y|x)}{\entropy{p_{\phi}(\theta)} - \entropy{p_{\phi}(\theta|x,y)}} \\
                     & = \expectation{p_{\phi}(\theta)}{\entropy{p_{\phi}(y|x)}-\entropy{p_{\phi}(y|x,\theta)}}
    .
\end{align*}
Notably BALD is often used even when updating is not Bayesian, for example when using Monte Carlo dropout in a neural network \citep{gal2017deep}.

\section{The shortfalls of BALD}
\label{sec:bald_shortfalls}

To establish the need for a new approach to Bayesian active learning, we highlight that BALD can be poorly suited to the prediction-oriented settings that constitute much of machine learning.
We explain that this stems from the mismatch that can exist between parameter uncertainty and predictive uncertainty.
We also highlight that targeting predictive uncertainty requires reasoning about what inputs we want to make predictions on, which BALD does not do.

\subsection{Focusing on prediction}

In statistics it is common for the model parameters to be valued in their own right \citep{beck1977parameter,blei2003latent,fisher1925statistical}.
But in many machine-learning contexts, particularly the supervised settings where BALD is typically applied, the parameters are only valued insofar as they serve a prediction-oriented goal.
We often, for example, seek the parameters that maximise the model's predictive performance on a test data distribution \citep{hastie2009elements}.
This frequentist notion of success often remains our motivation even if we use a Bayesian approach for either data acquisition or learning, or for both \citep{komaki1996asymptotic,snelson2005sparse}.

\subsection{Not all information is equal}

In some models, such as linear models, parameters and predictions are tightly coupled.
This means that a reduction in parameter uncertainty typically yields a wholesale reduction in predictive uncertainty \citep{chaloner1995bayesian}.
But more generally the coupling can be loose.
Deep neural networks, for instance, can have substantial redundancy in their parameters \citep{belkin2019reconciling}, while Bayesian nonparametric models can be thought of as having an infinite number of parameters \citep{hjort2010bayesian}.
When the coupling is loose, parameter uncertainty can be reduced without a corresponding reduction in predictive uncertainty on inputs of interest.
In fact it is possible to gain an infinite amount of parameter information while seeing an arbitrarily small reduction in predictive uncertainty.

\begin{example}
    \label{example:gp}
    Consider a supervised-learning problem where $x,y\in\mathbb{R}$ and we use a model consisting of a Gaussian likelihood function, $p(y|x,\theta) = \mathcal{N}(\theta(x), 1)$, and a zero-mean Gaussian-process prior, $\theta \sim \mathrm{GP}(0,k)$, with covariance function $k(x,x') = \exp(-(x-x')^2)$.
    Suppose we are interested in making predictions in the interval $x_* \in [0,1]$.
    Now consider gathering observations at the input locations $M,2M,\dots,M^2$ for some $M \in \mathbb{N}^+$.
    In the limit $M\to\infty$, BALD converges to infinity while the EIG in the prediction of interest, $\theta(x_*)$, converges to zero:
    \begin{align*}
        \lim_{M\to \infty} \mathrm{BALD}((M,2M,\dots,M^2))               & = \infty \\
        \lim_{M \to \infty} \mathrm{EIG}_{\theta(x_*)}((M,2M,\dots,M^2)) & = 0.
    \end{align*}
\end{example}
See \Cref{sec:app:example_proof} for the proof.
This example is a concrete demonstration that a high BALD score need not coincide with any reduction in the predictive uncertainty of interest, $\mathrm{EIG}_{\theta(x_*)}$.
If the aim is to predict, then maximising BALD is not guaranteed to help to any extent whatsoever.

\subsection{BALD has no notion of an input distribution}
\label{sec:input_dist}

In order to reason about what information is relevant to prediction, we need some notion of the inputs on which we want to make predictions.
Without this we have no mechanism to ensure the model we learn is well-suited to the task we care about.
Our model could be highly effective on inputs from one region of input space but useless for typical samples from an input distribution of interest.

Appreciating the need to account for which inputs might arise at test time, it becomes clear why BALD can be problematic.
BALD focuses on the model parameters in isolation, with no explicit connection to prediction.
As such, it does not account for the distribution over inputs.

\subsection{Real-world data can exacerbate this problem}

BALD can be particularly problematic in the very settings that often motivate active learning: those where we have access to a large pool of unlabelled inputs whose relevance to some task of interest varies widely.
In contrast with the carefully curated datasets often used in basic research, real-world data is often drawn from many sources of varying fidelity and relation to the task.
Pools of web-scraped audio, images and text are canonical examples of this.
Active learning ought to help deal with the mess by identifying only the most useful inputs to label.
But BALD can in fact be worse than random acquisition in these settings, targeting obscure data that is not helpful for prediction.

The experiment presented in \Cref{fig:bald_pool_size} highlights this flaw.
As we increase the size of the pool that BALD is maximised over, inputs of greater obscurity become more likely to be included in the pool, and BALD produces worse and worse predictive accuracy.
This result is corroborated by the work of \citet{karamcheti2021mind}.
Focusing on visual-question-answering tasks, they found that BALD failed to outperform random acquisition when using uncurated pools, and that a substantial amount of curation was required before this shortfall could be overturned.

\subsection{Failure can occur without distribution shift}

\begin{figure}[t]
    \centering
    \includegraphics[height=\figureheight]{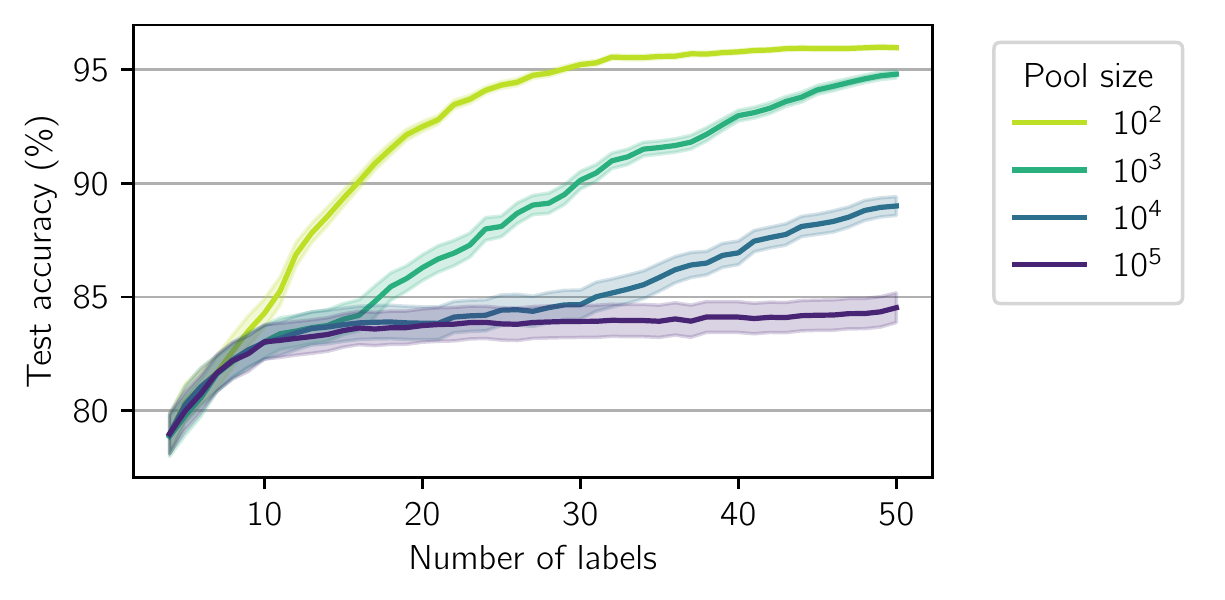}
    \caption{
        BALD can fail catastrophically on big pools.
        A bigger pool typically contains more inputs with low density under the data-generating distribution.
        Often these inputs are of low relevance if the aim is to maximise expected predictive performance.
        BALD can nevertheless favour these inputs.
        See \Cref{fig:heatmaps} for intuition and \Cref{sec:2d_data} for details.
    }
    \label{fig:bald_pool_size}
\end{figure}

It might be tempting to just think of this problem with BALD as being analogous to the issues caused by train-test input-distribution shifts elsewhere in machine learning.
But the problem is more deep-rooted than this: BALD has no notion of any input distribution in the first place.
This is why increasing the size of the pool can induce failures as in \Cref{fig:bald_pool_size}, without any distribution shift or changes to the distribution that the pool inputs are drawn from.
Distribution shift can cause additional problems for BALD, as some of the results in \Cref{sec:experiments} show.
But it is by no means a necessary condition for failure to occur.

\subsection{Filtering heuristics are not a general solution}
\label{sec:filtering}

We might suppose we could just discard irrelevant data before deploying BALD.
But this filtering process would require us to be able to determine each input's relevance at the outset of training, which is impractical in many cases.
Even if we have access to a target input distribution, this on its own can be insufficient for judging relevance to a task of interest.
A candidate input could have relatively low density under the target distribution but nevertheless share high-level features with a target input, such that the two inputs' labels are highly mutually informative.
With high-dimensional inputs, it can also be surprisingly difficult to identify unrepresentative inputs purely through their density~\citep{nalisnick2018deep}.
Rather than trying to design an auxiliary process to mitigate BALD's problematic behaviour, we seek an acquisition function that can automatically determine what is relevant.

\section{Expected predictive information gain}
\label{sec:method}

Motivated by BALD's weakness in prediction-oriented settings, we return to the framework of Bayesian experimental design that underlies BALD, and derive an acquisition function that we call the expected predictive information gain (EPIG).
Whereas BALD targets a reduction in parameter uncertainty, EPIG directly targets a reduction in predictive uncertainty on inputs of interest.

To reason about reducing predictive uncertainty, we need an explicit notion of the predictions we want to make with our model.
We therefore introduce a target input distribution, $\ptarg(x_*)$, and define our goal to be the confident prediction of the labels, $y_*$, associated with samples $x_* \sim \ptarg(x_*)$.

To derive EPIG we first consider the information gain in $y_*$ that results from conditioning on new data, $(x, y)$:
\begin{align*}
    \mathrm{IG}_{y_*}(x,y,x_*) = \entropy{p_\phi(y_*|x_*)} - \entropy{p_\phi(y_*|x_*,x,y)}
    ,
\end{align*}
where $p_\phi(y_*|x_*,x,y)=\expectation{p_\phi(\theta|x,y)}{p_\phi(y_*|x_*,\theta)}$.
Note that this is a function of $x_*$ as well as $x$ and $y$.
Next we take an expectation over both the random target input, $x_*$, and the unknown label, $y$:
\begin{align*}
    \mathrm{EPIG}(x) = \expectation{\ptarg(x_*) p_\phi(y|x)}{\mathrm{IG}_{y_*}(x,y,x_*)}
    .
\end{align*}
Thus we see that EPIG is the expected reduction in predictive uncertainty at a randomly sampled target input, $x_*$.

There are other interpretations too.
EPIG is the mutual information between $(x_*,y_*)$ and $y$ given $x$, $\mutualinfo{(x_*,y_*);y|x}$:
\begin{align}
    \mathrm{EPIG}(x) = \expectation{\ptarg(x_*) p_\phi(y,y_*|x,x_*)}{\log \frac{p_\phi(y,y_*|x,x_*)}{p_\phi(y|x)p_\phi(y_*|x_*)}}
    .
    \label{eq:epig_mi}
\end{align}
This is equivalent to $\expectation{\ptarg(x_*)}{\mutualinfo{y;y_*|x,x_*}}$, the expected mutual information between $y$ and $y_*$ given $x$ and $x_*$, which can be written as an expected KL divergence between $p_\phi(y,y_*|x,x_*)$ and $p_\phi(y|x)p_\phi(y_*|x_*)$: $\mathrm{EPIG}(x)=$
\begin{align}
    \expectation{\ptarg(x_*)}{\kldivergence{p_\phi(y,y_*|x,x_*)}{p_\phi(y|x)p_\phi(y_*|x_*)}}
    .
    \label{eq:epig_kl}
\end{align}

We can also take a frequentist perspective.
In classification settings EPIG is equal (up to a constant) to the negative expected generalisation error under a cross-entropy loss:
\begin{align}
    \label{eq:gen_loss}
    \mathrm{EPIG}(x) = \expectation{\ptarg(x_*) p_\phi(y,y_*|x,x_*)}{\log p_\phi(y_*|x_*,x,y)} + c
    ,
\end{align}
where $c$ is a constant and we have used the fact that $\entropy{p_\phi(y_*|x_*)}$ is constant with respect to $x$.
Maximising EPIG can therefore be thought of as seeking to minimise the expected generalisation error after acquisition.

\subsection{Sampling target inputs}
\label{sec:sampling_x_star}

EPIG involves an expectation with respect to a target input distribution, $\ptarg(x_*)$.
In practice we estimate this expectation by Monte Carlo and so require a sampling mechanism.

In many active-learning settings an input distribution is implied by the existence of a pool of unlabelled inputs.
There are cases where we know (or are happy to assume) the pool has been sampled from $\ptarg(x_*)$.
Alternatively we might be forced to assume this is the case: perhaps we know the pool is not sampled from $\ptarg(x_*)$ but lack access to anything better.
In these cases we can simply subsample from the pool to obtain samples of $x_*$.
Empirically we find that this can work well relative to acquisition with BALD (\Cref{sec:experiments}).

Another important case is where we have access to samples from $\ptarg(x_*)$ but we cannot label them.
Limits on the ability to acquire labels might arise due to privacy-related and other ethical concerns, geographical restrictions, the complexity of the labelling process for some inputs, or the presence of commercially sensitive information in some inputs.
At the same time there might be a pool of inputs for which we have no labelling restrictions.
In a case like this we can estimate EPIG using samples from $\ptarg(x_*)$ while using only the pool as a source of candidates for labelling.
Thus we can target information gain in predictions on samples from $\ptarg(x_*)$ without labelling those samples themselves.

A further scenario that we might encounter is a classification problem where the pool is representative of the target class-conditional input distribution but not the target marginal class distribution: that is, $\ppool(x_*|y_*)=\ptarg(x_*|y_*)$ but $\ppool(y_*)\neq\ptarg(y_*)$.
The pool might, for example, consist of uncurated web-scraped inputs from many more classes than those we care about.
In this scenario it can often be the case that we know or can reasonably approximate the distribution over classes that we are targeting, $\ptarg(y_*)$.
With this we can approximately sample from $\ptarg(x_*)$ using a combination of our model, $p_{\phi}(y|x)$, and the pool.
We first note that
\begin{align*}
    \ppool(x_*|y_*) = \frac{\ppool(x_*) \ppool(y_*|x_*)}{\int \ppool(x)\ppool(y=y_*|x) dx}
    .
\end{align*}
Then, using the fact that $\ppool(x_*|y_*) = \ptarg(x_*|y_*)$, we get
\begin{align*}
    \ptarg(x_*) & = \sum_{y_*\in\mathcal{Y}} \ptarg(y_*) \ptarg(x_*|y_*)                                                                            \\
                & = \ppool(x_*) \sum_{y_*\in\mathcal{Y}} \frac{\ptarg(y_*) \ppool(y_*|x_*)}{\int p_{\mathrm{pool}}(x)p_{\mathrm{pool}}(y=y_*|x) dx} \\
                & \approx \ppool(x_*) \sum_{y_*\in\mathcal{Y}} \frac{\ptarg(y_*) p_\phi(y_*|x_*)}{\frac{1}{N} \sum_{x\in\pooldata} p_\phi(y=y_*|x)} \\
                & = \ppool(x_*) w(x_*)
    ,
\end{align*}
where we have approximated $\ppool(y_*|x_*)$ with our model.
Now we can approximately sample from $\ptarg(x_*)$ by subsampling inputs from the pool using a categorical distribution with probabilities $w(x_*)/N$.

\subsection{Estimation}
\label{sec:estimation}

The best way to estimate EPIG depends on the task and model of interest.
In the empirical evaluations in this paper we focus on classification problems and use models whose marginal and joint predictive distributions are not known in closed form.
This leads us to use $\mathrm{EPIG}(x) \approx$
\begin{equation}
    \label{eq:epig_fast_estimator}
    \frac{1}{M} \sum_{j=1}^{M} \kldivergence{\hat{p}_\phi(y,y_*|x,x_*^j)}{\hat{p}_\phi(y|x)\hat{p}_\phi(y_*|x_*^j)}
    ,
\end{equation}
where $x_*^j \sim \ptarg(x_*)$ and the instances of $\hat{p}$ denote Monte Carlo approximations of the predictive distributions in \Cref{eq:marg_pred_posterior,eq:joint_pred_posterior}.
Specifically we define
\begin{align*}
    \hat{p}_{\phi}(y|x)             & = \frac{1}{K} \sum_{i=1}^K p_{\phi}(y|x,\theta_i)
    \\
    \hat{p}_{\phi}(y_1,y_2|x_1,x_2) & = \frac{1}{K} \sum_{i=1}^K p_{\phi}(y_1,y_2|x_1,x_2,\theta_i)
    ,
\end{align*}
where $\theta_i \sim p_\phi(\theta)$.
Classification is an instance of where the required expectation over $y$ and $y_*$ can be computed analytically, such that our only required estimation is from marginalisations over $\theta$.

If we cannot integrate over $y$ and $y_*$ analytically, we can revert to nested Monte Carlo estimation \citep{rainforth2018nesting}.
For this we first note that, using \Cref{eq:joint_pred_posterior}, we can sample $y,y_* \sim p_\phi(y,y_*|x,x_*)$ exactly by drawing a $\theta$ and then a $y$ and $y_*$ conditioned on this $\theta$.
By also drawing samples for $\theta$, we can then construct the estimator $\mathrm{EPIG}(x) \approx$
\begin{align}
    \label{eq:epig_sample_estimator}
    \frac{1}{M} \sum_{j=1}^{M} \log \frac{
        K \sum_{i=1}^K p_\phi(y^{j}|x,\theta_i)p_\phi(y^{j}_*|x_*^{j},\theta_i)
    }{
        \sum_{i=1}^K p_\phi(y^{j}|x,\theta_i) \sum_{i=1}^K p_\phi(y^{j}_*|x_*^{j},\theta_i)
    }
    ,
\end{align}
where $x_*^{j} \sim \ptarg(x_*)$, $y^{j},y_*^{j}\sim p_\phi(y,y_*|x,x_*^{j})$ and $\theta_i \sim p_\phi(\theta)$.
Subject to some weak assumptions on $p_{\phi}$, this estimator converges as $K,M\to\infty$ \citep{rainforth2018nesting}.

The EPIG estimators in \Cref{eq:epig_fast_estimator,eq:epig_sample_estimator} each have a total computational cost of $O(MK)$.
This is comparable to BALD estimation for regression problems.
But it can be more expensive than BALD estimation for classification problems: BALD can be collapsed to a non-nested Monte Carlo estimation for an $O(K)$ cost, but EPIG cannot.

Other possible estimation schemes include a variational approach inspired by \citet{foster2019variational}.
This is too expensive to be practically applicable in the settings we consider but could be useful elsewhere.
See \Cref{sec:epig_estimation} for details.
\section{Experiments}
\label{sec:experiments}

For consistency with existing work on active learning for prediction, our empirical evaluation of EPIG focuses on classification problems.
Code for reproducing our results is available at \shorturl{github.com/fbickfordsmith/epig}.

\begin{figure}[t]
    \centering
    \includegraphics[height=\figureheight]{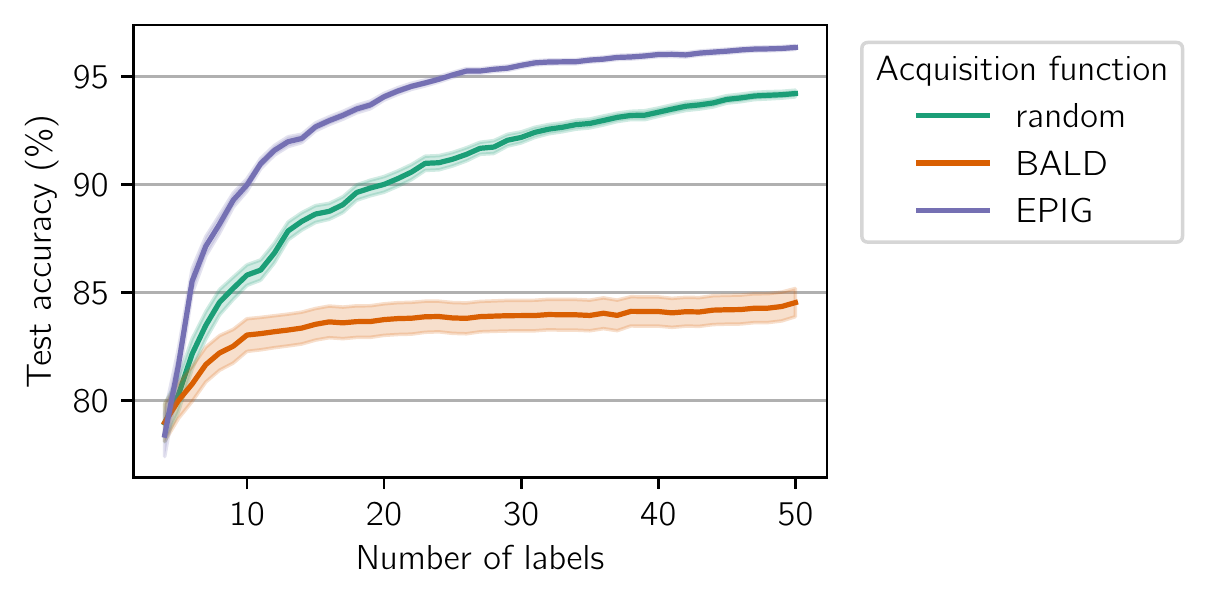}
    \caption{
        In contrast with BALD, EPIG deals effectively with a big pool ($10^5$ unlabelled inputs).
        BALD is overwhelmingly counterproductive relative to random acquisition.
        See \Cref{fig:heatmaps,fig:bald_pool_size} for intuition and \Cref{sec:2d_data} for details.
    }
    \label{fig:2d_data}
\end{figure}

\subsection{Synthetic data (\Cref{fig:heatmaps,fig:bald_pool_size,fig:2d_data})}
\label{sec:2d_data}

First we demonstrate the difference between BALD and EPIG in a setting that is easy to visualise and understand: binary classification with two-dimensional inputs.

\paragraph{Data}

The first input distribution of interest, denoted $p_1(x)$ in \Cref{fig:heatmaps}, is a bivariate Student's $t$ distribution with $\nu=5$ degrees of freedom, location $\mu=[0,0]$ and scale matrix $\Sigma=0.8I$.
The second distribution, denoted $p_2(x)$ in \Cref{fig:heatmaps}, is a scaled and shifted version of the first, with parameters $\nu=5$, $\mu\approx[0.8,0.9]$ and $\Sigma=0.4I$.
This serves to illustrate in \Cref{fig:heatmaps} how EPIG's value changes with the target input distribution; it is not used elsewhere.
The conditional label distribution is defined as $p(y=1|x)=\Phi(20 (\tanh(2x_{[1]}) - x_{[2]}))$, where $x_{[i]}$ denotes the component of input $x$ in dimension $i$, and $\Phi$ is the cumulative distribution function of the standard normal distribution.
For the training data in \Cref{fig:heatmaps}, we sample ten input-label pairs, $\traindata=\{(x_i,y_i)\}_{i=1}^{10}$, where $x_i, y_i \sim p(y|x)p_1(x)$.
Likewise we sample $\testdata=\{(x_i,y_i)\}_{i=1}^{10,000}$ for evaluating the model's performance in active learning.

\paragraph{Model and training}

We use a model with a probit likelihood function, $p_\phi(y=1|x,\theta)=\Phi(\theta(x))$, where $\Phi$ is defined as above, and a Gaussian-process prior, $\theta \sim \mathrm{GP}(0,k)$, where $k(x,x')=10\cdot\exp\left(-\sfrac{\|x-x'\|^2}{2}\right)$.
The posterior over latent-function values cannot be computed exactly so we optimise an approximation to it using variational inference \citep{hensman2015scalable}.
To do this we run 10,000 steps of full-batch gradient descent using a learning rate of 0.005 and a momentum factor of 0.95.

\paragraph{Active learning}

We initialise the training dataset, $\traindata$, with two randomly sampled inputs from each class.
Thereafter we run the active-learning loop described in \Cref{sec:active_learning} until a budget of 50 labels is used up.
We acquire data using three acquisition functions: random, BALD and EPIG.
Random acquisition involves sampling uniformly from the pool without replacement.
We estimate BALD using \Cref{eq:bald_classification_estimator} and EPIG using \Cref{eq:epig_fast_estimator}, in both cases drawing 5,000 samples from the model's approximate posterior.
For EPIG we sample $x_*\sim p_1(x_*)$.
After each time the model is trained, we evaluate its predictive accuracy on $\testdata$ as defined above.
Using a different random-number-generator seed each time, we run active learning with each acquisition function 100 times.
We report the test accuracy (mean $\pm$ standard error) as a function of the size of $\mathcal{D}_\mathrm{train}$.

\paragraph{Discussion}

\Cref{fig:2d_data} shows a striking gap between BALD and EPIG in active learning.
\Cref{fig:heatmaps,fig:bald_pool_size} provide some intuition about the underlying cause of this disparity: BALD has a tendency to acquire labels at the extrema of the input space, regardless of their relevance to the predictive task of interest.

\begin{figure*}[t]
    \centering
    \includegraphics[height=\figureheight]{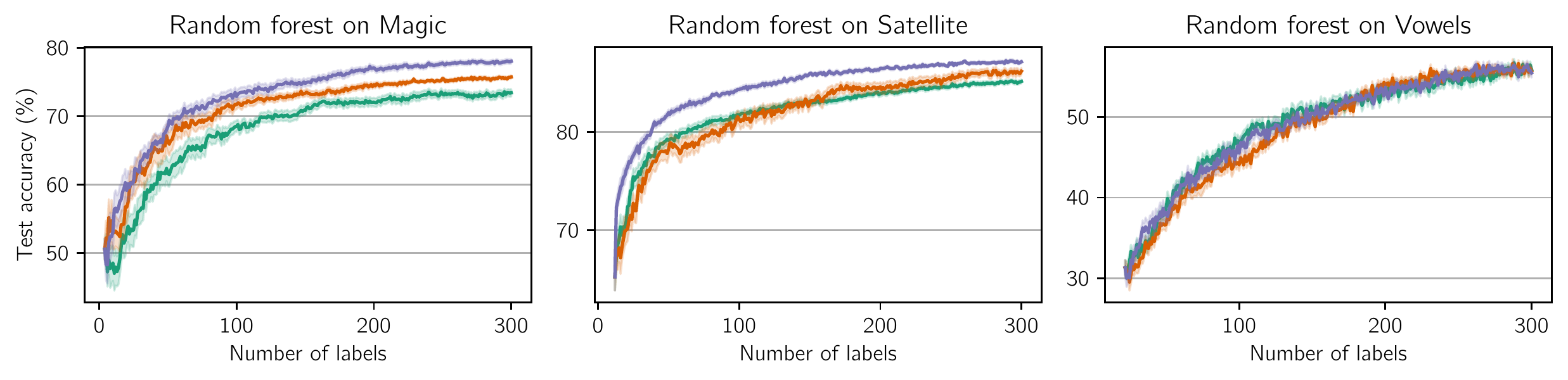}
    \includegraphics[height=\figureheight]{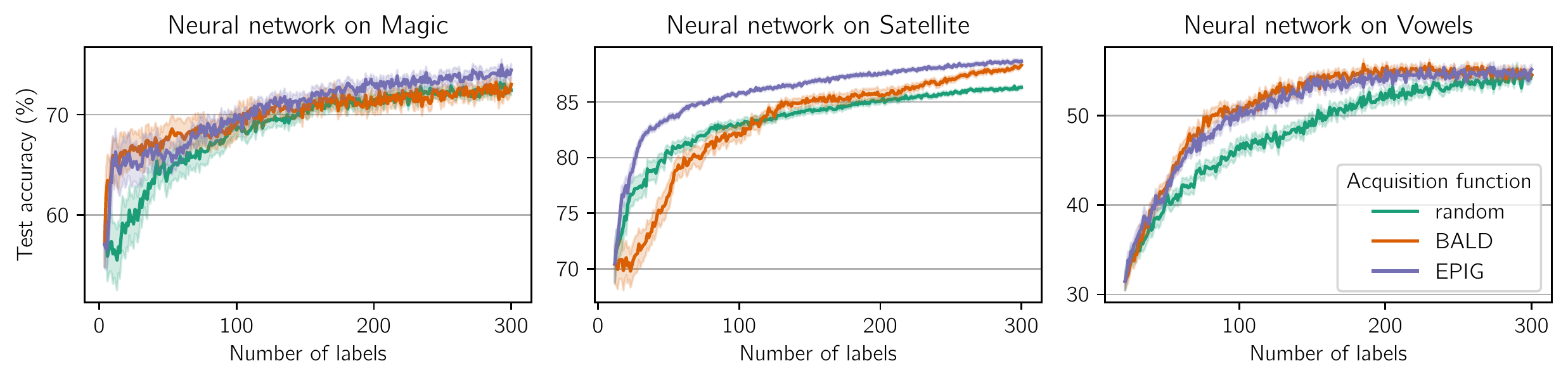}
    \caption{
        EPIG outperforms or matches BALD across three standard classification problems from the UCI machine-learning repository (Magic, Satellite and Vowels) and two models (random forest and neural network).
        See \Cref{sec:uci} for details.
    }
    \label{fig:uci}
\end{figure*}

\subsection{UCI data (\Cref{fig:uci})}
\label{sec:uci}

Next we compare BALD and EPIG in a broader range of settings.
We use problems drawn from a repository maintained at UC Irvine (UCI; \citealp{dua2017uci}), which has been widely used as a data source in past work on Bayesian methods \citep{gal2016dropout,lakshminarayanan2017simple,sun2018differentiable,zhang2018noisy}.
The problems we use vary in terms of the number of classes, the input dimension and any divergence between the pool and target data distributions.
We assume knowledge of $\ptarg(x_*)$ when estimating EPIG but note that this assumption has little significance if $\ppool(x)$ and $\ptarg(x_*)$ match, which is true for two out of the three problems.

\paragraph{Data}

We use three classification datasets from the UCI repository, each with a different number of classes, $C$, and input dimension, $D$: Magic ($C=2$, $D=11$), Satellite ($C=6$, $D=36$) and Vowels ($C=11$, $D=10$).
The inputs are telescope readings in Magic, satellite images in Satellite and speech recordings in Vowels.
Magic is interesting because it serves as a natural instance of a mismatch between pool and target distributions (see \Cref{sec:uci_appendix}).

\paragraph{Models and training}

We use two different models.
The first is a random forest \citep{breiman2001random}.
To emphasise that EPIG works with an off-the-shelf setup, we use the Scikit-learn \citep{pedregosa2011scikitlearn} implementation with its default parameters.
The second model is a dropout-enabled fully connected neural network with three hidden layers and a softmax output layer.
A dropout rate of 0.1 is used during both training and testing.
Training the neural network consists of running up to 50,000 steps of full-batch gradient descent using a learning rate of 0.1.
We use a loss function consisting of the negative log likelihood (NLL) of the training data combined with an $l_2$ regulariser (with coefficient 0.0001) on the model parameters.
To mitigate overfitting we use early stopping: we track the model's NLL on a small validation set (approximately 20\% of the size of the training-label budget) and stop training if this NLL does not decrease for more than 10,000 consecutive steps.
We then restore the model parameters to the configuration that achieved the lowest validation-set NLL.

\paragraph{Active learning}

We use largely the same setup as described in \Cref{sec:2d_data}.
Here the label budget is 300 and we run active learning 20 times with different seeds.
We use the same BALD and EPIG estimators as before, treating each tree in the random forest as a different $\theta$ value, and treating each stochastic forward pass through the neural network (we compute 100 of them) as corresponding to a different $\theta$ value.
To estimate EPIG we sample $x_*$ from a set of inputs designed to be representative of $\ptarg(x_*)$.

\paragraph{Discussion}

\Cref{fig:uci} shows EPIG performing convincingly better than BALD in some cases while matching it in others.
These results provide broader validation of EPIG, complementing the results in \Cref{fig:2d_data}.

\begin{figure*}[t]
    \centering
    \includegraphics[height=\figureheight]{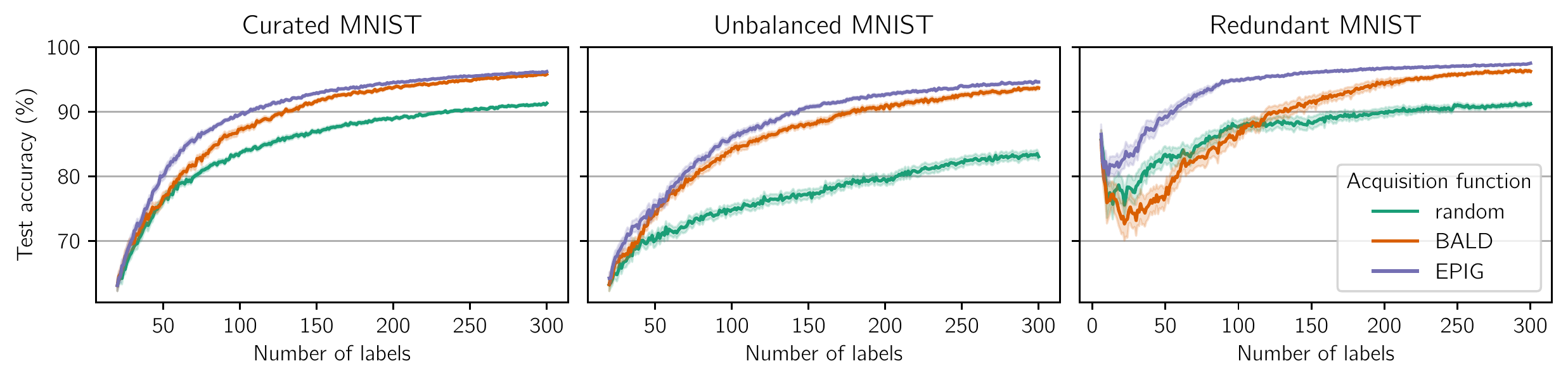}
    \caption{
        EPIG outperforms BALD across three image-classification settings.
        Curated MNIST reflects the data often used in academic research.
        The pool and target input distributions, $\ppool(x)$ and $\ptarg(x_*)$ match; the marginal class distributions, $\ppool(y)$ and $\ptarg(y_*)$, are uniform.
        Unbalanced MNIST is a step closer to real-world data.
        While $\ptarg(y_*)$ remains uniform, $\ppool(y)$ is non-uniform: the pool contains more inputs from some classes than others.
        Redundant MNIST simulates a separate practical problem.
        Whereas $\ptarg(y_*)$ only has nonzero mass on two classes of interest, $\ppool(y)$ has substantial mass across all classes.
        See \Cref{sec:mnist} for details.
    }
    \label{fig:mnist_standard}
\end{figure*}

\begin{figure*}[t]
    \centering
    \includegraphics[height=\figureheight]{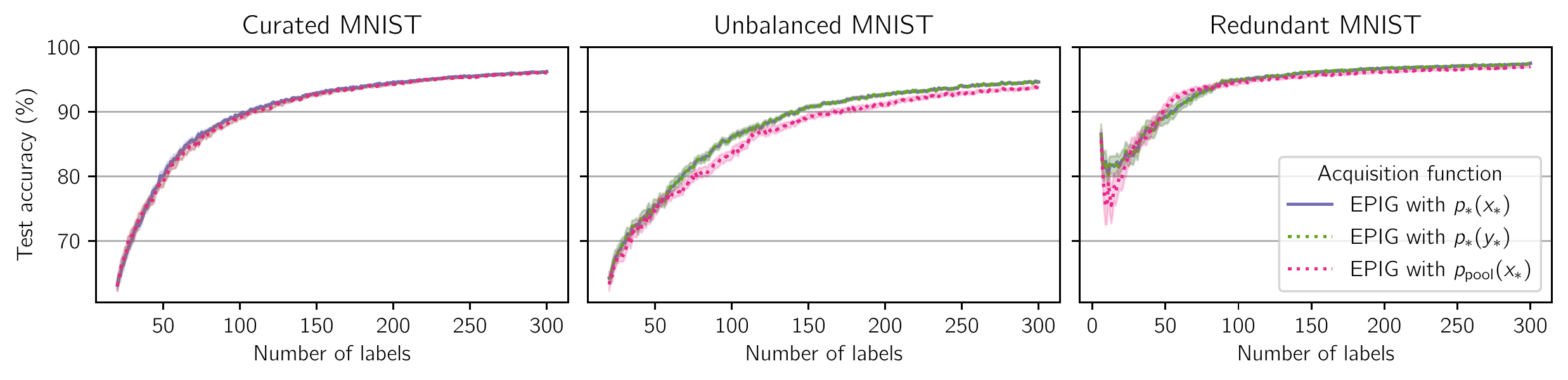}
    \caption{
        Even without knowledge of the target input distribution, $\ptarg(x_*)$, EPIG retains its strong performance on Curated MNIST, Unbalanced MNIST and Redundant MNIST.
        ``EPIG with $\ptarg(x_*)$'' assumes exact samples from $\ptarg(x_*)$, as in \Cref{fig:mnist_standard}.
        ``EPIG with $\ptarg(y_*)$'' corresponds to using the approximate sampling scheme outlined in \Cref{sec:sampling_x_star}, using knowledge of $\ptarg(y_*)$.
        ``EPIG with $\ppool(x_*)$'' corresponds to using samples from the pool as a proxy for $\ptarg(x_*)$.
        See \Cref{sec:mnist} for details.
    }
    \label{fig:mnist_ablation}
\end{figure*}

\subsection{MNIST data (\Cref{fig:mnist_standard,fig:mnist_ablation,fig:mnist_extra_baselines})}
\label{sec:mnist}

Finally we evaluate BALD and EPIG in settings intended to capture challenges that occur when applying deep neural networks to high-dimensional inputs.
Our starting point is the MNIST dataset \citep{lecun1998gradientbased}, in which each input is an image of a handwritten number between 0 and 9.
This dataset has been widely used in related work on Bayesian active learning with deep neural networks \citep{beluch2018power,gal2017deep,jeon2020thompsonbald,kirsch2019batchbald,kirsch2022stochastic,lee2019baldvae,tran2019bayesian}.
We construct three settings based on this dataset, each corresponding to a different practical scenario: Curated MNIST, Unbalanced MNIST and Redundant MNIST.

As well as investigating how BALD and EPIG perform across these settings, we seek to understand the effect on EPIG of varying the amount of knowledge we have of the target data distribution, $\ptarg(x_*)$.
To this end we assume we know this for one set of runs (\Cref{fig:mnist_standard}) and then relax this assumption for another set (\Cref{fig:mnist_ablation}).

\paragraph{Data}

Curated MNIST is intended to reflect the data often used in academic machine-learning research.
The pool and target class distributions, $\ppool(y)$ and $\ptarg(y_*)$, are both uniform over all 10 classes.
In terms of class distributions, this effectively represents a worst-case scenario for active learning relative to random acquisition.
Given matching class-conditional input distributions, namely $\ppool(x_*|y_*)=\ptarg(x_*|y_*)$, uniformly sampling from the pool input distribution, $\ppool(x)$, is equivalent to uniformly sampling from the target input distribution, $\ptarg(x_*)$.
Thus random acquisition is a strong baseline in this setting.

Unbalanced MNIST is a step closer to real-world data.
We might expect $\ptarg(y_*)$ to be uniform---that is, the task of interest might involve classifying examples in equal proportion from each class---but it could be difficult to curate a pool that is similarly uniform in its class distribution.
To reflect this we consider a case with a non-uniform $\ppool(y)$: classes 0 to 4 each have probability $\sfrac{1}{55}$ and classes 5 to 9 each have probability $\sfrac{10}{55}$.

Redundant MNIST captures a separate practical problem that occurs, for instance, when using web-scraped data.
The pool might contain inputs from many more classes than we want to focus on in the predictive task of interest.
To simulate this we suppose that the task involves classifying just images of 1s and 7s, occurring in equal proportion---that is, $\ptarg(y_*)$ places probability mass of $\sfrac{1}{2}$ on class 1, $\sfrac{1}{2}$ on class 7, and 0 on all other classes---while $\ppool(y)$ is uniform over all 10 classes.
If the acquisition function selects an input from a class other than 1 and 7, the labelling function produces a ``neither'' label.
Thus we have three-way classification during training: 1 vs 7 vs neither.

\paragraph{Model and training}

For both runs we use the same dropout-enabled convolutional neural network as used by \citet{kirsch2019batchbald}.
The dropout rate here is 0.5.
Training is similar to as described in \Cref{sec:uci}, except that the learning rate is 0.01 and early stopping triggers after 5,000 consecutive steps of non-decreasing validation-set NLL.

\paragraph{Active learning}

Initially we retain the setup described in \Cref{sec:uci}, with $\ptarg(x_*)$ known (\Cref{fig:mnist_standard}).
Then we investigate the sensitivity of EPIG to removing full access to $\ptarg(x_*)$, focusing on two different settings (\Cref{fig:mnist_ablation}).
In one we assume knowledge of $\ptarg(y_*)$ and use the resampling technique discussed in \Cref{sec:sampling_x_star}.
In the other we simply sample target inputs from the pool: $x_* \sim \ppool(x_*)$.

\paragraph{Discussion}

\Cref{fig:mnist_standard} shows EPIG again outperforming BALD and random across all three dataset variants when given access to $\ptarg(x_*)$.
(EPIG additionally beats predictive entropy \citep{settles2008analysis} and BADGE \citep{ash20deep}, acquisition functions commonly studied in the active-learning literature, as shown in \Cref{sec:app:extra_results}.)
EPIG's advantage over BALD is appreciable on Curated MNIST and Unbalanced MNIST.
But it is emphatic on Redundant MNIST.
This suggests EPIG is particularly useful when working with highly diverse pools.

\Cref{fig:mnist_ablation} shows the even more impressive result that EPIG retains its strong performance even when no access to $\ptarg(x_*)$ is assumed.
We thus see that EPIG provides a good degree of robustness in its performance to the level of knowledge about the target data distribution.
\section{Related work}

The idea of using the EIG to quantify the utility of data was introduced by \citet{lindley1956measure} and has a long history of use in experimental design \citep{chaloner1995bayesian,rainforth2023modern}.
The framework of Bayesian experimental design has many applications outside active learning, and in these applications the model parameters are commonly the quantity of interest---Bayesian optimisation \citep{hennig2012entropy,hernandezlobato2016predictive,villemonteix2009informational} being a notable exception.
The EIG in the parameters is thus often a natural acquisition function.

The EIG in the parameters was suggested as an acquisition function for active learning by \citet{mackay1992evidence,mackay1992information}, who called it the total information gain.
It was popularised as BALD by \citet{houlsby2011bayesian}, while its use with deep neural networks was demonstrated by \citet{gal2017deep}.
BALD has been widely used elsewhere in cases where the model's predictions, not the model parameters, are the true objects of interest \citep{atighehchian2020bayesian,beluch2018power,jeon2020thompsonbald,kirsch2019batchbald,kirsch2022stochastic,lee2019baldvae,munjal2022towards,pinsler2019bayesian,shen2018deep,siddhant2018deep,tran2019bayesian}.

Maximising the information gathered about a quantity other than the model parameters has been proposed a number of times as an approach to active learning.
Perhaps most relevant to our work, \citet{mackay1992evidence,mackay1992information} introduced an acquisition function called the mean marginal information gain.
Based on a Gaussian approximation of the posterior over the model parameters, it measures the average information gain in the predictions made on a fixed set of inputs.
Though it has since received surprisingly little attention in the literature, it was discussed by \citet{huszar2013scoring} and later used by \citet{wang2021beyond} to evaluate the quality of predictive-posterior correlations.
Seeking information gain on a fixed set of inputs---in contrast with the input distribution considered by EPIG---is a transductive approach to active learning \citep{vapnik1982estimation,yu2006active}.

Aside from the work of \citet{mackay1992evidence,mackay1992information}, there are numerous prediction-oriented methods \citep{afrabandpey2019human,chapelle2005active,cohn1993neural,cohn1996active,daee2017knowledge,donmez2008optimizing,evans2015estimating,filstroff2021targeted,krause2008near,seo2000gaussian,sundin2018improving,sundin2019active,tan2021diversity,yu2006active,zhao2021bayesian,zhao2021efficient,zhao2021uncertainty,zhu2003combining}.
Many of these, with notable examples including the work of \citet{cohn1996active} and \citet{krause2008near}, are tied to a particular model class or approximation scheme and so lack EPIG's generality.

There is an additional limitation associated with techniques based on the idea, due to \citet{roy2001toward}, of measuring the expected loss reduction that would result from updating the model on a given input-label pair.
These techniques often require updating the model within the computation of the acquisition function, which can be extremely expensive.
Despite a strong conceptual connection to the acquisition function proposed by \citet{roy2001toward}, EPIG allows a significantly lower computational cost: its information-theoretic formulation allows us to derive an estimator that does not require nested model updating.
\section{Conclusion}

We have demonstrated that BALD, a widely used acquisition function for Bayesian active learning, can be suboptimal.
While much of machine learning focuses on prediction, BALD targets information gain in a model's parameters in isolation and so can seek labels that have limited relevance to the predictions of interest.
Motivated by this, we have proposed EPIG, an acquisition function that targets information gain in terms of predictions.
Our results show EPIG outperforming BALD across a number of data settings (low- and high-dimensional inputs, varying degrees of divergence between the pool and target data distributions, and varying degrees of knowledge of the target distribution) and across multiple different models.
This suggests EPIG can serve as a compelling drop-in replacement for BALD, with particular scope for performance gains when using large, diverse pools of unlabelled data.
\section*{Acknowledgements}

We thank Arnaud Doucet, Mike Osborne, Jannik Kossen, Jan Brauner, Joost van Amersfoort, Florentin Coeurdoux and the anonymous reviewers of this paper for their feedback.
Freddie Bickford Smith and Andreas Kirsch are supported by the EPSRC Centre for Doctoral Training in Autonomous Intelligent Machines and Systems (EP/L015897/1, EP/S024050/1).

\bibliography{references}

\appendix
\onecolumn

\section{Proof for \Cref{example:gp}}
\label{sec:app:example_proof}

Let $\mathbf{X} = (M, 2M, \dots, M^2)$ denote a collection of inputs, $\mathbf{y} = (y_1,y_2,\dots,y_{M})$ denote their labels, and $\mathbf{f} = (\theta(M),\theta(2M),\dots,\theta(M^2))$ denote the values of the Gaussian process at $\mathbf{X}$.
The conditional distribution of $\mathbf{y}$ given $\mathbf{f}$ and the marginal distribution of $\mathbf{y}$ are both multivariate Gaussian.
Their covariance matrices are
\begin{align*}
    \mathrm{Cov}(\mathbf{y} | \mathbf{f}, \mathbf{X}) & = I_{M}                                     \\
    \mathrm{Cov}(\mathbf{y} | \mathbf{X})             & = \begin{pmatrix}
                                                              2         & e^{-M^2} & e^{-4M^2} & \dots  \\
                                                              e^{-M^2}  & 2        & e^{-M^2}  & \dots  \\
                                                              e^{-4M^2} & e^{-M^2} & 2         & \dots  \\
                                                              \vdots    & \vdots   & \vdots    & \ddots
                                                          \end{pmatrix}.
\end{align*}
We now use two facts: that $\mathrm{BALD}(\mathbf{X}) = \expectation{\mathbf{f}}{\entropy{\mathbf{y}} - \entropy{\mathbf{y} | \mathbf{f}}}$; and that for a multivariate Gaussian random variable, $Z \sim \mathcal{N}(\mu, \Sigma)$, the entropy is $\entropy{Z} = \frac{1}{2} \log \det 2 \pi e \Sigma$ \citep{cover2005elements}.
Combining these with the simple form of $\mathrm{Cov}(\mathbf{y} | \mathbf{f}, \mathbf{X})$, in particular its independence from $\mathbf{f}$, gives
\begin{align}  \label{eq:app:det}
    \mathrm{BALD}(\mathbf{X}) = \frac{1}{2}\log \begin{vmatrix}
                                                    2         & e^{-M^2} & e^{-4M^2} & \dots  \\
                                                    e^{-M^2}  & 2        & e^{-M^2}  & \dots  \\
                                                    e^{-4M^2} & e^{-M^2} & 2         & \dots  \\
                                                    \vdots    & \vdots   & \vdots    & \ddots
                                                \end{vmatrix} := \frac{1}{2}\log\det \Omega_M.
\end{align}
To control this determinant we apply the Gershgorin circle theorem \citep{horn2012matrix}, which states that the eigenvalues of $\Omega_M$ in \Cref{eq:app:det} lie within the interval $[2 - \varepsilon_M, 2 + \varepsilon_M]$, where
\begin{align*}
    \varepsilon_M = \sum_{i\ne j}^{M} e^{-M^2|i-j|^2} \le M^2 e^{-M^2} \to 0 \text{ as } M \to \infty.
\end{align*}
We therefore have
\begin{align*}
    (1 - \varepsilon_M/2)^M \le \frac{\det \Omega_M}{2^M} \le (1 + \varepsilon_M/2)^M.
\end{align*}
Taking the limit as $M\to\infty$ we have
\begin{align*}
    \log \left((1 - \varepsilon_M/2)^M\right) = M \log (1 - \varepsilon_M/2) = -\frac{1}{2}M^3e^{-M^2} + O\left(M^6e^{-2M^2}\right) \to 0 \text{ as } M \to \infty,
\end{align*}
with a similar result for $(1 + \varepsilon_M/2)$.
From this we deduce
\begin{align*}
    \mathrm{BALD}(\mathbf{X}) = \frac{1}{2}\log \det \Omega_M \to \frac{1}{2} \log 2^{M} \to \infty \text{ as } M \to \infty.
\end{align*}

Next we turn to $\mathrm{EIG}_{\theta(x_*)}(\mathbf{X})$.
We have the covariance matrix
\begin{align*}
    \mathrm{Cov}(\theta(x_*), \mathbf{y}  | \mathbf{X}) = \begin{pmatrix}
                                                              1                 & e^{-|x_* - M|^2} & e^{-|x_* - 2M|^2} & \dots  \\
                                                              e^{-|x_* - M|^2}  & 2                & e^{-M^2}          & \dots  \\
                                                              e^{-|x_* - 2M|^2} & e^{-M^2}         & 2                 & \dots  \\
                                                              \vdots            & \vdots           & \vdots            & \ddots
                                                          \end{pmatrix}.
\end{align*}
Now consider a set of random variables, $\theta(x_*)'$ and $\mathbf{y}'$, that have the same marginal distributions as $\theta(x_*)$ and $\mathbf{y}$ respectively but are independent of each other.
Thus $\theta(x_*)'$ and $\mathbf{y}'$ are jointly Gaussian with covariance matrix
\begin{align*}
    \mathrm{Cov}(\theta(x_*)', \mathbf{y}' | \mathbf{X}) = \begin{pmatrix}
                                                               1      & 0        & 0        & \dots  \\
                                                               0      & 2        & e^{-M^2} & \dots  \\
                                                               0      & e^{-M^2} & 2        & \dots  \\
                                                               \vdots & \vdots   & \vdots   & \ddots
                                                           \end{pmatrix}.
\end{align*}
Using the fact that $\mathrm{EIG}_{\theta(x_*)}(\mathbf{X}) = \mutualinfo{\theta(x_*); \mathbf{y} | \mathbf{X}} = \entropy{(\theta(x_*)',\mathbf{y}')} - \entropy{(\theta(x_*),\mathbf{y})}$, along with the formula used above for the entropy of a multivariate Gaussian random variable, we can write
\begin{align*}
    \mathrm{EIG}_{\theta(x_*)}(\mathbf{X}) = \frac{1}{2} \log \det \mathrm{Cov}(\theta(x_*)', \mathbf{y}'  | \mathbf{X}) - \frac{1}{2} \log \det \mathrm{Cov}(\theta(x_*), \mathbf{y}  | \mathbf{X}).
\end{align*}
Noting that you can remove a factor from any row of a matrix as a prefactor on the determinant, for both matrices we remove the factors of 2 from each row except the first:
\begin{align*}
    \mathrm{EIG}_{\theta(x_*)}(\mathbf{X}) = & \frac{1}{2} \log 2^{M-1} \begin{vmatrix}
                                                                            1      & 0          & 0          & \dots  \\
                                                                            0      & 1          & e^{-M^2}/2 & \dots  \\
                                                                            0      & e^{-M^2}/2 & 1          & \dots  \\
                                                                            \vdots & \vdots     & \vdots     & \ddots
                                                                        \end{vmatrix}                            \\
                                             & -\frac{1}{2} \log 2^{M-1} \begin{vmatrix}
                                                                             1                   & e^{-|x_* - M|^2} & e^{-|x_* - 2M|^2} & \dots  \\
                                                                             e^{-|x_* - M|^2}/2  & 1                & e^{-M^2}/2        & \dots  \\
                                                                             e^{-|x_* - 2M|^2}/2 & e^{-M^2}/2       & 1                 & \dots  \\
                                                                             \vdots              & \vdots           & \vdots            & \ddots
                                                                         \end{vmatrix}
    .
\end{align*}
As the $\log 2^{M-1}$ terms then cancel out,
\begin{align}  \label{eq:diff_det2}
    \mathrm{EIG}_{\theta(x_*)}(\mathbf{X}) =  \frac{1}{2} \log \begin{vmatrix}
                                                                   1      & 0          & 0          & \dots  \\
                                                                   0      & 1          & e^{-M^2}/2 & \dots  \\
                                                                   0      & e^{-M^2}/2 & 1          & \dots  \\
                                                                   \vdots & \vdots     & \vdots     & \ddots
                                                               \end{vmatrix} - \frac{1}{2} \log \begin{vmatrix}
                                                                                                    1                   & e^{-|x_* - M|^2} & e^{-|x_* - 2M|^2} & \dots  \\
                                                                                                    e^{-|x_* - M|^2}/2  & 1                & e^{-M^2}/2        & \dots  \\
                                                                                                    e^{-|x_* - 2M|^2}/2 & e^{-M^2}/2       & 1                 & \dots  \\
                                                                                                    \vdots              & \vdots           & \vdots            & \ddots
                                                                                                \end{vmatrix}.
\end{align}
Finally we apply the Gershgorin circle theorem again, along with the fact that $|x_*-iM| \ge |M-1|$ for $i=1,2,\dots$.
We conclude that the eigenvalues of both matrices in \Cref{eq:diff_det2} lie within the interval $[1 - \varepsilon_M, 1 + \varepsilon_M]$, where
\begin{align*}
    \varepsilon_M = \sum_{i\ne j}^{M} e^{-(x_*-|i-j|M)^2} \le M^2 e^{-|M-1|^2} \to 0 \text{ as } M \to \infty.
\end{align*}
Therefore both determinants are bounded below by $(1 - \varepsilon_M)^M$ and above by $(1 + \varepsilon_M)^M$.
Taking the limit as $M\to\infty$ we have
\begin{align*}
    \log \left((1 - \varepsilon_M)^M\right) = M \log (1 - \varepsilon_M) = -M^3e^{-|M-1|^2} + O\left(M^6e^{-2|M-1|^2}\right) \to 0 \text{ as } M \to \infty,
\end{align*}
with a similar result for $(1 + \varepsilon_M)$.
Thus both determinants in \Cref{eq:diff_det2} converge to 1 as $M\to\infty$.
From this we conclude
\begin{align*}
    \mathrm{EIG}_{\theta(x_*)}(\mathbf{X}) \to 0 \text{ as } M \to \infty.
\end{align*}
\section{BALD derivation}
\label{sec:bald_derivation}

The information gain in $\theta$ due to $(x,y)$ is the reduction in Shannon entropy in $\theta$ that results from observing $(x,y)$:
\begin{align*}
    \mathrm{IG}_{\theta}(x,y) = \entropy{p_\phi(\theta)} - \entropy{p_\phi(\theta|x,y)}
    ,
\end{align*}
where $p_\phi(\theta|x,y) \propto p_\phi(y|x,\theta)p_\phi(\theta)$ is the posterior after updating on $(x,y)$.

\newpage

Since $y$ is a random variable, we compute the expected information gain in $\theta$, known as the BALD score.
To do this we use the model's marginal predictive distribution, $p_\phi(y|x)=\expectation{p_\phi(\theta)}{p_\phi(y|x,\theta)}$, to simulate the labels we might observe:
\begin{align}
    \mathrm{BALD}(x)
     & =
    \expectation{p_\phi(y|x)}{
        \mathrm{IG}_\theta(x,y)
    }
    \nonumber
    \\
     & =
    \expectation{p_\phi(y|x)}{
        \entropy{p_\phi(\theta)} - \entropy{p_\phi(\theta|x,y)}
    }
    \nonumber
    \\
     & =
    \expectation{p_\phi(y|x)}{
        -\expectation{p_\phi(\theta)}{\log p_\phi(\theta)}
        +\expectation{p_\phi(\theta|x,y)}{\log p_\phi(\theta|x,y)}
    }
    \nonumber
    \\
     & =
    \expectation{p_\phi(\theta)p_\phi(y|x,\theta)}{
        \log\frac{p_\phi(\theta|x,y)}{p_\phi(\theta)}
    }
    \nonumber
    \\
     & =
    \expectation{p_\phi(\theta)p_\phi(y|x,\theta)}{
        \log\frac{p_\phi(y|x,\theta)}{p_\phi(y|x)}
    }
    \nonumber
    \\
     & =
    \expectation{p_\phi(\theta)p_\phi(y|x,\theta)}{
        -\log p_\phi(y|x) + \log p_\phi(y|x,\theta)
    }
    \nonumber
    \\
     & =
    \expectation{p_\phi(\theta)}{
        -\expectation{p_\phi(y|x)}{\log p_\phi(y|x)}
        +\expectation{p_\phi(y|x,\theta)}{\log p_\phi(y|x,\theta)}
    }
    \nonumber
    \\
     & =
    \expectation{p_\phi(\theta)}{
        \entropy{p_\phi(y|x)} - \entropy{p_\phi(y|x,\theta)}
    }
    \label{eq:bald_entropy_diff}
    .
\end{align}
\section{BALD estimation}
\label{sec:bald_estimation}

In general we can estimate BALD using nested Monte Carlo \citep{rainforth2018nesting}:
\begin{align*}
    \mathrm{BALD}(x)
     & =
    \expectation{p_\phi(\theta)}{
        -\expectation{p_\phi(y|x)}{\log p_\phi(y|x)}
        +\expectation{p_\phi(y|x,\theta)}{\log p_\phi(y|x,\theta)}
    }
    \\
     & \approx
    \frac{1}{M} \sum_{j=1}^M
    -\log\left(\frac{1}{K}\sum_{i=1}^K p_\phi(y_j|x,\theta_i)\right)
    + \log p_\phi(y_j|x,\theta_j)
    ,
\end{align*}
where $\theta_i \sim p_{\phi}(\theta)$, $(\theta_j, y_j)\sim p_\phi(\theta)p_\phi(y|x,\theta)$.
Special cases allow us to use computationally cheaper estimators.

\subsection{Categorical predictive distribution}

When $y$ and $y_*$ are discrete we can write
\begin{align*}
    \mathrm{BALD}(x)
     & =
    \expectation{p_\phi(\theta)}{
        -\expectation{p_\phi(y|x)}{\log p_\phi(y|x)}
        +\expectation{p_\phi(y|x,\theta)}{\log p_\phi(y|x,\theta)}
    }
    \\
     & =
    -\expectation{p_\phi(y|x)}{\log p_\phi(y|x)}
    +\expectation{p_\phi(\theta)p_\phi(y|x,\theta)}{\log p_\phi(y|x,\theta)}
    \\
     & =
    -\sum_{y\in\mathcal{Y}} p_\phi(y|x) \log p_\phi(y|x)
    +\expectation{p_\phi(\theta)}{\sum_{y\in\mathcal{Y}} p_\phi(y|x,\theta) \log p_\phi(y|x,\theta)}
    .
\end{align*}
This can be estimated using samples, $\theta_i\sim p_\phi(\theta)$ \citep{houlsby2014efficient}:
\begin{align}
    \label{eq:bald_classification_estimator}
    \mathrm{BALD}(x)
     & \approx
    -\sum_{y\in\mathcal{Y}} \hat{p}_\phi(y|x) \log \hat{p}_\phi(y|x)
    +\frac{1}{K} \sum_{i=1}^K \sum_{y\in\mathcal{Y}} p_\phi(y|x,\theta_i) \log p_\phi(y|x,\theta_i)
    ,
\end{align}
where
\begin{align*}
    \hat{p}_\phi(y|x) = \frac{1}{K}\sum_{i=1}^K p_\phi(y|x,\theta_i)
    .
\end{align*}

\subsection{Gaussian predictive distribution}

Suppose we have a model whose likelihood function, $p_\phi(y|x,\theta)$, and predictive distribution, $p_\phi(y|x)$, are Gaussian.
Then, using \Cref{eq:bald_entropy_diff} along with knowledge of the entropy of a Gaussian \citep{cover2005elements}, we have
\begin{align*}
    \mathrm{BALD}(x)
    = \frac{1}{2} \log 2 \pi e \mathbb{V}[p_\phi(y|x)] - \expectation{p_\phi(\theta)}{\frac{1}{2} \log 2 \pi e \mathbb{V}[p_\phi(y|x,\theta)]}
    = \frac{1}{2} \left(\log \mathbb{V}[p_\phi(y|x)] - \expectation{p_\phi(\theta)}{\log \mathbb{V}[p_\phi(y|x,\theta)]} \right)
    .
\end{align*}
Relatedly \citet{houlsby2011bayesian} identified a closed-form approximation of BALD for the particular case of using a probit likelihood function, a Gaussian-process prior and a Gaussian approximation to the predictive distribution.
\section{EPIG derivation}
\label{sec:epig_derivation}

The information gain in $y_*$ due to $(x,y)$ is the reduction in Shannon entropy in $y_*$ that results from observing $(x,y)$:
\begin{align*}
    \mathrm{IG}_{y_*}(x,y,x_*) = \entropy{p_{\phi}(y_*|x_*)} - \entropy{p_{\phi}(y_*|x_*,x,y)}
    ,
\end{align*}
where $p_\phi(y_*|x_*,x,y)=\expectation{p_\phi(\theta|x,y)}{p_\phi(y_*|x_*,\theta)}$.

Computing an expectation over both $y$ and $x_*$ gives the expected predictive information gain (EPIG):
\begin{align*}
    \mathrm{EPIG}(x)
     & =
    \expectation{\ptarg(x_*) p_\phi(y|x)}{\mathrm{IG}_{y_*}(x,y,x_*)}
    \\
     & =
    \expectation{\ptarg(x_*) p_\phi(y|x)}{\entropy{p_\phi(y_*|x_*)} - \entropy{p_\phi(y_*|x,y,x_*)}}
    \\
     & =
    \expectation{\ptarg(x_*) p_\phi(y|x)}{
        -\expectation{p_\phi(y_*|x_*)}{\log p_\phi(y_*|x_*)}
        +\expectation{p_\phi(y_*|x,y,x_*)}{\log p_\phi(y_*|x,y,x_*)}
    }
    \\
     & =
    \expectation{\ptarg(x_*) p_\phi(y,y_*|x,x_*)}{
        \log\frac{p_\phi(y_*|x,y,x_*)}{p_\phi(y_*|x_*)}
    }
    \\
     & =
    \expectation{\ptarg(x_*) p_\phi(y,y_*|x,x_*)}{
        \log\frac{p_\phi(y|x)p_\phi(y_*|x,y,x_*)}{p_\phi(y|x)p_\phi(y_*|x_*)}
    }
    \\
     & =
    \expectation{\ptarg(x_*) p_\phi(y,y_*|x,x_*)}{
        \log\frac{p_\phi(y,y_*|x,x_*)}{p_\phi(y|x)p_\phi(y_*|x_*)}
    }
    \\
     & =
    \expectation{\ptarg(x_*)}{
        \mutualinfo{y;y_*|x,x_*}
    }
    \\
     & =
    \expectation{\ptarg(x_*)}{
        \kldivergence{p_\phi(y,y_*|x,x_*)}{p_\phi(y|x)p_\phi(y_*|x_*)}
    }.
\end{align*}
\section{EPIG estimation}
\label{sec:epig_estimation}

While in general we can use \Cref{eq:epig_sample_estimator} to estimate EPIG, special cases allow computationally cheaper estimators.

\subsection{Categorical predictive distribution}

When $y$ and $y_*$ are discrete we can write
\begin{align*}
    \mathrm{EPIG}(x)
     & =
    \expectation{\ptarg(x_*)}{
        \kldivergence{p_\phi(y,y_*|x,x_*)}{p_\phi(y|x)p_\phi(y_*|x_*)}
    }
    \\
     & =
    \expectation{\ptarg(x_*)}{
        \sum_{y\in\mathcal{Y}} \sum_{y_*\in\mathcal{Y}} p_\phi(y,y_*|x,x_*) \log \frac{p_\phi(y,y_*|x,x_*)}{p_\phi(y|x)p_\phi(y_*|x_*)}
    }
    .
\end{align*}
This can be estimated using samples, $\theta_i\sim p_\phi(\theta)$ and $x_*^j\sim \ptarg(x_*)$:
\begin{align*}
    \mathrm{EPIG}(x)
     & \approx
    \frac{1}{M} \sum_{j=1}^M \sum_{y\in\mathcal{Y}} \sum_{y_*\in\mathcal{Y}}
    \hat{p}_\phi(y,y_*|x,x_*^j) \log \frac{\hat{p}_\phi(y,y_*|x,x_*^j)}{\hat{p}_\phi(y|x)\hat{p}_\phi(y_*|x_*^j)}
    ,
\end{align*}
where
\begin{align*}
    \hat{p}_\phi(y,y_*|x,x_*^j)
     & =
    \frac{1}{K}\sum_{i=1}^K p_\phi(y|x,\theta_i)p_\phi(y_*|x_*^j,\theta_i)
    \\
    \hat{p}_\phi(y|x)
     & =
    \frac{1}{K}\sum_{i=1}^K p_\phi(y|x,\theta_i)
    \\
    \hat{p}_\phi(y_*|x_*^j)
     & =
    \frac{1}{K}\sum_{i=1}^K p_\phi(y_*|x_*^j,\theta_i)
    .
\end{align*}

\subsection{Gaussian predictive distribution}
\label{sec:epig_estimation_gaussian}

Consider a joint predictive distribution that is multivariate Gaussian with mean vector $\mu$ and covariance matrix $\Sigma$:
\begin{align*}
    p_\phi(y,y_*|x,x_*)
    =
    \mathcal{N}(\mu, \Sigma)
    =
    \mathcal{N}
    \left(
    \mu,
    \begin{bmatrix}
            \mathrm{cov}(x,x)   & \mathrm{cov}(x,x_*)   \\
            \mathrm{cov}(x,x_*) & \mathrm{cov}(x_*,x_*)
        \end{bmatrix}
    \right)
    .
\end{align*}
In this setting the mutual information between $y$ and $y_*$ given $x$ and $x_*$ is a closed-form function of $\Sigma$:
\begin{align*}
    \mutualinfo{y;y_*|x,x_*}
     & =
    \entropy{p_\phi(y|x)} + \entropy{p_\phi(y_*|x_*)} - \entropy{p_\phi(y,y_*|x,x_*)}
    \\
     & =
    \frac{1}{2} \log 2 \pi e \mathbb{V}[p_\phi(y|x)] + \frac{1}{2} \log 2 \pi e \mathbb{V}[p_\phi(y_*|x_*)]
    - \frac{1}{2} \log \det 2 \pi e \Sigma
    \\
     & =
    \frac{1}{2} \log \frac{
        \mathbb{V}[p_\phi(y|x)] \mathbb{V}[p_\phi(y_*|x_*)]
    }{
        \det\Sigma
    }    \\
     & =
    \frac{1}{2} \log \frac{
        \mathrm{cov}(x,x) \mathrm{cov}(x_*,x_*)
    }{
        \det\Sigma
    }    \\
     & =
    \frac{1}{2} \log \frac{
        \mathrm{cov}(x,x) \mathrm{cov}(x_*,x_*)
    }{
        \mathrm{cov}(x,x) \mathrm{cov}(x_*,x_*) - \mathrm{cov}(x,x_*)^2
    }
    .
\end{align*}
We can estimate EPIG using samples, $x_*^j\sim \ptarg(x_*)$:
\begin{align*}
    \mathrm{EPIG}(x)
    =\expectation{\ptarg(x_*)}{\mutualinfo{y;y_*|x,x_*}}
    \approx
    \frac{1}{M} \sum_{j=1}^M \mutualinfo{y;y_*|x,x_*^j}
    =
    \frac{1}{2M} \sum_{j=1}^M \log \frac{
        \mathrm{cov}(x,x) \mathrm{cov}(x_*^j,x_*^j)
    }{
        \mathrm{cov}(x,x) \mathrm{cov}(x_*^j,x_*^j) - \mathrm{cov}(x,x_*^j)^2
    }
    .
\end{align*}

\subsection{Connection with \citet{foster2019variational}}
\label{sec:app:vboed_connection}

\citet{foster2019variational} primarily considered variational estimation of the expected information gain.
Since the joint density, $p_\phi(y,y_*|x,x_*)$, that appears in EPIG is often not known in closed form, EPIG estimation broadly falls under the ``implicit likelihood'' category of methods considered in that paper.
Here we focus on showing how the ``posterior'' or Barber-Agakov bound \citep{barber2003im} from this earlier work applies to EPIG estimation.
We first recall \Cref{eq:gen_loss},
\begin{align*}
    \mathrm{EPIG}(x) = \expectation{\ptarg(x_*) p_\phi(y,y_*|x,x_*)}{\log p_\phi(y_*|x_*,x,y)} + \entropy{p_\phi(y_*|x_*)}
    ,
\end{align*}
and the observation that $c = \entropy{p_\phi(y_*|x_*)}$ does not depend upon $x$ and hence can be neglected when choosing between candidate inputs.
By Gibbs's inequality we must have
\begin{align*}
    \mathrm{EPIG}(x) \ge \expectation{\ptarg(x_*) p_\phi(y,y_*|x,x_*)}{\log q(y_*|x_*,x,y)} + \entropy{p_\phi(y_*|x_*)}
\end{align*}
for any distribution $q$.
We can now consider a variational family, $q_\psi(y_*|x_*,x,y)$, and a maximisation over the variational parameter, $\psi$:
\begin{align*}
    \mathrm{EPIG}(x) \ge \sup_\psi \expectation{\ptarg(x_*) p_\phi(y,y_*|x,x_*)}{\log q_\psi(y_*|x_*,x,y)} + \entropy{p_\phi(y_*|x_*)}
    .
\end{align*}
A practical implication of this bound is that we could estimate EPIG by learning an auxiliary network, $q_\psi(y_*|x_*,x,y)$, using data simulated from the model to make one-step-ahead predictions.
That is, $q_\psi$ is trained to make predictions at $x_*$, incorporating the knowledge of the hypothetical acquisition $(x,y)$.
For our purposes, training such an auxiliary network at each acquisition is prohibitively expensive.
But this approach might be valuable in other applications of EPIG.
\section{Dataset construction}

\subsection{UCI data}
\label{sec:uci_appendix}

For each dataset we start by taking the base dataset, $\basedata$, from the UCI repository.
Satellite and Vowels have predefined test datasets, $\testdata$.
In contrast, Magic does not have a predefined train-test split.
It is stated in Magic's documentation that one of the classes is underrepresented in the dataset relative to real-world data (Magic is a simulated dataset).
Whereas classes 0 and 1 respectively constitute 65\% and 35\% of the dataset, it is stated that class 1 constitutes the majority of cases in reality (the exact split is not stated; we assume 75\% for class 1).
We therefore uniformly sample 30\% of $\basedata$ to form a test base dataset, $\basedata'$; then we set $\basedata\gets\basedata\setminus\basedata'$; then we make $\testdata$ by removing input-label pairs from $\basedata'$ such that class 1 constitutes 75\% of the subset.
With the test set defined, we proceed to sample two disjoint subsets of $\basedata$ such that their class proportions match those of $\basedata$: a pool set, $\pooldata$, whose size varies between datasets, and a validation set, $\mathcal{D}_\mathrm{val}$, of 60 input-label pairs.
Regardless of the class proportions of $\basedata$, we always use an initial training dataset, $\mathcal{D}_\mathrm{init}$, of 2 input-label pairs per class, sampled from $\basedata$.
Finally we sample a representative set of inputs, $\mathcal{D}_*$, whose class proportions match those of $\testdata$.

\subsection{MNIST data}
\label{sec:mnist_appendix}

Implementing each setting starts by using the standard MNIST training data (60,000 input-label pairs) as the base dataset, $\basedata$, and the standard MNIST testing data (10,000 input-label pairs) as the test base dataset, $\basedata'$.
For Redundant MNIST we make $\testdata$ by removing input-label pairs from $\basedata'$ such that only classes 1 and 7 remain.
Otherwise we set $\testdata=\basedata'$.
Next we construct the pool set, $\pooldata$.
For Curated MNIST and Redundant MNIST we sample 4,000 inputs per class from $\basedata'$.
For Unbalanced MNIST we sample 400 inputs per class for classes 0-4 and 4,000 inputs per class for classes 5-9.
After this we make the initial training dataset, $\mathcal{D}_\mathrm{init}$.
For Curated MNIST and Unbalanced MNIST we sample 2 input-label pairs per class from $\basedata$.
For Redundant MNIST we sample 2 input-label pairs from class 1, 2 input-label pairs from class 7 and 1 input-label pair per class from 2 randomly selected classes other than 1 and 7.
Next, the validation set, $\mathcal{D}_\mathrm{val}$.
For all settings this comprises 60 input-label pairs such that the class proportions match those used to form $\pooldata$.
Finally we sample a representative set of inputs, $\mathcal{D}_*$, whose class proportions match those of $\testdata$.
\section{Extra results}
\label{sec:app:extra_results}

\begin{figure*}[h!]
    \centering
    \includegraphics[height=\figureheight]{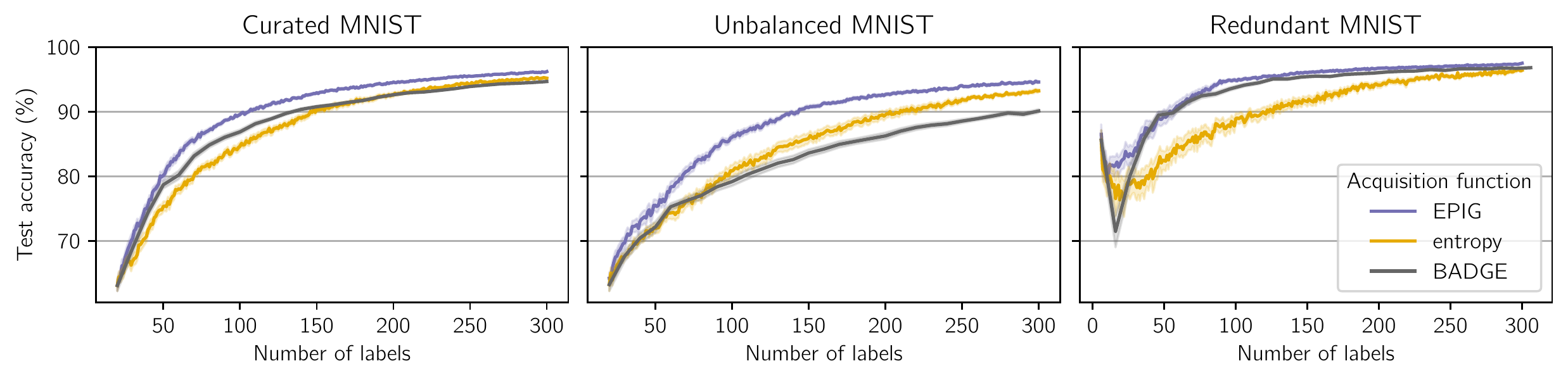}
    \caption{
        EPIG outperforms two acquisition functions popularly used as baselines in the active-learning literature.
        The first is the model's predictive entropy, $\entropy{p_\phi(y|x)}$ \citep{settles2008analysis}.
        The second is BADGE \citep{ash20deep}.
        Calculating BADGE involves computing a gradient-based embedding for each candidate input in the pool and then applying $k$-means++ initialisation \citep{arthur2007kmeansplusplus} in embedding space to select a diverse batch of inputs for labelling.
        We acquire 10 labels at a time with BADGE.
    }
    \label{fig:mnist_extra_baselines}
\end{figure*}

\end{document}